\documentclass[11pt]{article}

\usepackage[preprint]{acl}

\usepackage{times}
\usepackage{latexsym}
\usepackage{amsmath}

\usepackage{booktabs}
\usepackage[table]{xcolor}
\usepackage{colortbl}
\usepackage{makecell}
\usepackage[most]{tcolorbox}

\definecolor{vdus}{HTML}{E11D48}
\definecolor{vdusspill}{HTML}{FFF1F4}
\definecolor{vdresid}{HTML}{FFE4E6}
\definecolor{vdcn}{HTML}{047857}
\definecolor{vdcnspill}{HTML}{ECFDF5}
\definecolor{vdsame}{HTML}{ECFDF5}
\definecolor{vdgray}{HTML}{6A6A6A}
\definecolor{vdmllm}{HTML}{F59E0B}

\newcommand{\vdcell}[3]{\makecell[r]{%
  #1~{\scriptsize\textcolor{vdgray}{[#3\%]}}\\[-1pt]%
  {\scriptsize\textcolor{vdgray}{#2}}%
}}
\newcommand{\vdcellbest}[3]{\makecell[r]{%
  \textbf{#1}~{\scriptsize\textcolor{vdgray}{[#3\%]}}\\[-1pt]%
  {\scriptsize\textcolor{vdgray}{#2}}%
}}

\newtcbox{\pillUS}{
  on line, arc=3pt, outer arc=3pt, boxrule=0pt,
  colback=vdusspill, coltext=vdus,
  boxsep=0pt, left=4pt, right=4pt, top=0.5pt, bottom=0.5pt,
  fontupper=\scriptsize\bfseries\sffamily
}
\newtcbox{\pillCN}{
  on line, arc=3pt, outer arc=3pt, boxrule=0pt,
  colback=vdcnspill, coltext=vdcn,
  boxsep=0pt, left=4pt, right=4pt, top=0.5pt, bottom=0.5pt,
  fontupper=\scriptsize\bfseries\sffamily
}
\newcommand{\originus}{\pillUS{US}}
\newcommand{\origincn}{\pillCN{CN}}

\tcbset{
  promptbox/.style={
    colback=gray!4,
    colframe=black!70,
    boxrule=0.4pt,
    arc=2pt,
    left=8pt, right=8pt, top=6pt, bottom=6pt,
    fonttitle=\bfseries\small\raggedright,
    fontupper=\small,
    fontlower=\small,
    sharp corners=south,
    sidebyside,
    sidebyside align=top,
    sidebyside gap=10pt,
    segmentation style={solid, black!30},
    enhanced,
  },
  promptbox-single/.style={
    colback=gray!4,
    colframe=black!70,
    boxrule=0.4pt,
    arc=2pt,
    left=8pt, right=8pt, top=6pt, bottom=6pt,
    fonttitle=\bfseries\small,
    fontupper=\small,
    sharp corners=south,
    enhanced,
  }
}

\usepackage[T1]{fontenc}

\usepackage[utf8]{inputenc}

\usepackage{microtype}

\usepackage{inconsolata}

\usepackage{graphicx}
\usepackage{enumitem}

\usepackage{CJKutf8}
\newenvironment{cn}{\begin{CJK}{UTF8}{gbsn}}{\end{CJK}}

\newcommand{\eg}{\textit{e.g.,}}

\newcommand{\usnum}[1]{\textcolor{vdus}{\textbf{#1}}}
\newcommand{\cnnum}[1]{\textcolor{vdcn}{\textbf{#1}}}

\newcommand{\benchmark}{\textsc{VOIR DIRE}}

\title{Jury Duty: Calibration and Orientation Failures
\\ in MLLM-as-a-Judge Under Cultural Ambiguity}

\author{
 \textbf{Daniel Lee\thanks{\ These authors contributed equally.}\textsuperscript{1}},
 \textbf{Harsh Sharma\footnotemark[1]\textsuperscript{3}},
 \textbf{Eunkyu Park\textsuperscript{4}},
 \textbf{Pranav Venkit\textsuperscript{1}},
\\
 \textbf{Jeonghwan Kim\textsuperscript{5}},
 \textbf{Kah Mun Chia\textsuperscript{1}},
 \textbf{Andreas Vlachos\textsuperscript{2}},
 \textbf{Shafiq Joty\textsuperscript{1}}
\\
\\
 \textsuperscript{1}Salesforce AI Research,
 \textsuperscript{2}University of Cambridge,
 \\
 \textsuperscript{3}University of Colorado Boulder,
 \textsuperscript{4}Carnegie Mellon University,
 \textsuperscript{5}UIUC
\\
 \small{
   \textbf{Correspondence:} \href{mailto:d.lee2@salesforce.com}{d.lee2@salesforce.com}
 }
}

\begin{document}
\maketitle
\begin{abstract}
MLLM-as-a-Judge is conventionally validated by agreement with human annotations, but agreement against a culturally heterogeneous pool is pool-dependent: it silently conditions on whose judgments the aggregate represents. We introduce \benchmark{}, a multimodal benchmark of 626 culturally paired image--prompt items spanning U.S.\ and mainland Chinese contexts across food, fashion, and architecture, with annotator pools that are within-pool reliable ($\alpha = 0.86 / 0.74$) but cross-pool divergent on evaluation (Q1 $r = -0.12$). Across six MLLMs, the bias decomposes into two failures: a positivity-floor calibration failure (compressed scale use) and an orientation failure (default to one cultural norm). On this corpus, where contested items are sampled to split the two pools, the floor mechanically validates the more-permissive Chinese reading; persona prompting partially recovers calibration, but the orientation residual survives, evidence the tilt is not reducible to scale compression. Reference-pool in-context demonstrations deepen the orientation residual and inflate the high end rather than restoring use of the low end. Model origin adds a small additive tilt ($\sim 0.10$ MAE) that is approximately invariant under demonstration.
We recommend reporting alignment against each reference pool separately and treating cross-pool divergence as a judge property.
\end{abstract}
\section{Introduction} \label{1_introduction}
LLM-as-a-Judge~\cite{zheng2023judging}, along with multimodal adaptations such as MLLM-as-a-Judge~\cite{10.5555/3692070.3692324} and fine-tuned judge models~\cite{xu2026foundational,wang-etal-2025-direct-judgement,zhu2025judgelm, kim2024prometheus}, has become a widely used framework for evaluating open-ended generation, summarization, image generation, and other tasks where reference-based metrics are limited. These approaches are cost-effective and scalable~\cite{dubois2024lengthcontrolled}, and are conventionally validated by measuring agreement with human annotations or preferences~\cite{liu-etal-2023-g}.

\begin{figure*}[t]
        \centering       \includegraphics[width=.8\textwidth]{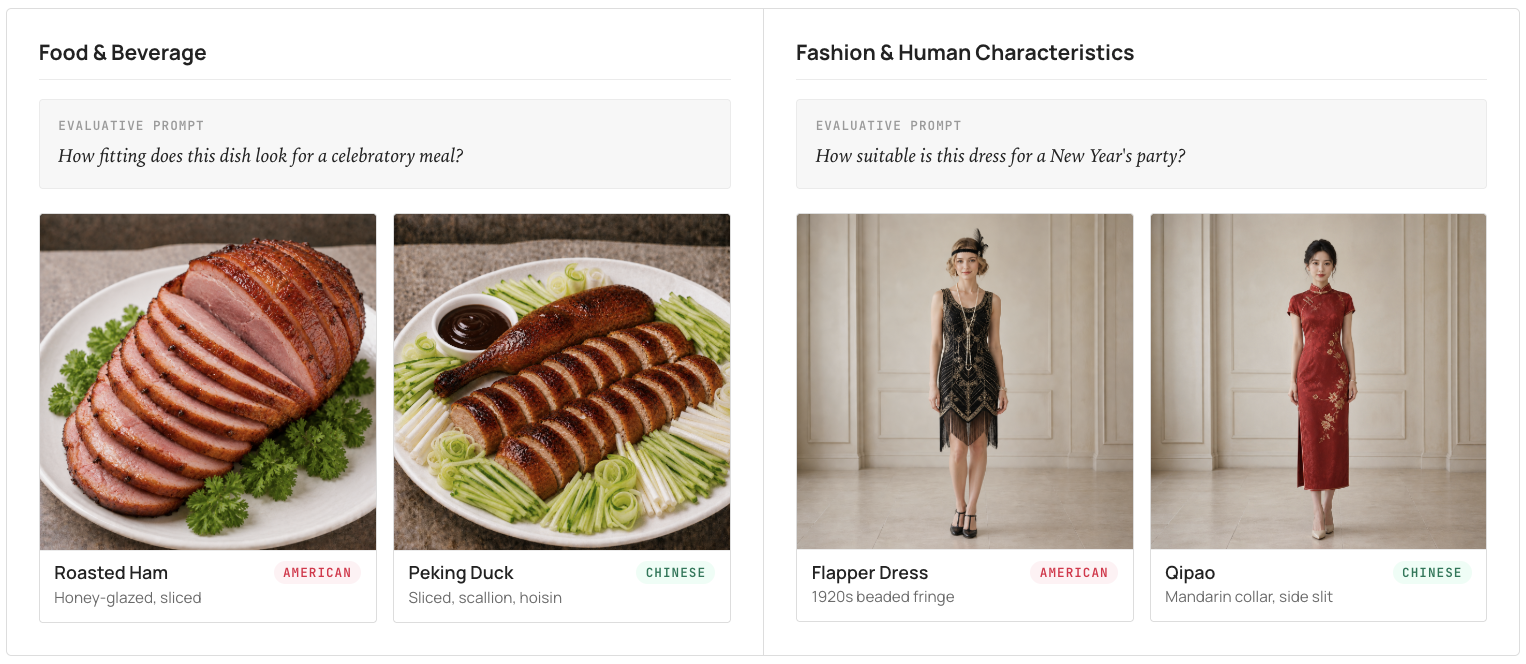}
    \caption{Illustrative concept-paired stimuli: each pair shares one evaluative prompt applied to a U.S.-aligned and a mainland-Chinese-aligned
  image across the Food and Beverage and Fashion and Human Characteristics categories.}
    \label{fig:fig1}
\end{figure*}

Validation by human agreement, however, does not specify whose judgments a model approximates~\cite{10.5555/3618408.3619652, durmus2024towards}. The distinction is consequential in preference-sensitive settings, where evaluations turn on plausibility, appropriateness, or desirability~\cite{ye2024flask} rather than verifiable correctness. Unlike tasks like code generation that is evaluated against unit tests~\cite{Chen2021EvaluatingLL}, such judgments admit multiple defensible answers, and cross-annotator differences may reflect distinct norms or values rather than error~\cite{pavlick-kwiatkowski-2019-inherent}.

Consider a culturally ambiguous case: U.S.\ contexts associate convenience-store food with low quality, while mainland Chinese contexts treat it as a source of fresh, reliable prepared meals. A judge evaluating such a scene draws on culturally situated expectations of plausibility and desirability, not just factual knowledge. Evaluations that appear culturally neutral are not: they reflect cultural assumptions the judge does not declare and the metric does not surface.

Prior work on cultural bias in multimodal models has largely documented misrepresentation in generation~\cite{naous-etal-2024-beer, liu2025culturevlmcharacterizingimprovingcultural}; whether such biases also shape evaluation, where model preferences are not visible as outputs but folded into scalar scores, has received less attention. The setting is structurally consequential: judge outputs propagate as training signal in RLAIF and judge-distilled benchmarks, and as ground truth in leaderboards, so cultural lean in a judge becomes cultural lean in a metric for the next training cycle. The concern is amplified for MLLMs developed in distinct regional markets (e.g., Claude, GPT, Llama in the U.S.; Qwen, Kimi, GLM in mainland China) and for multimodal evaluation, where culturally specific meanings are conveyed through visual cues~\cite{nayak-etal-2024-benchmarking, li-etal-2024-foodieqa, kim-etal-2025-tom}.

We investigate this through \benchmark{},\footnote{Named after the legal process of questioning prospective jurors to uncover hidden bias before trial.} a multimodal benchmark comparing MLLM-as-a-Judge alignment with culturally distinct human reference pools. Our contributions are threefold. \textbf{(1)} We release \benchmark{}, a benchmark of 626 culturally paired image--prompt items spanning U.S.\ and mainland Chinese contexts, with a construction protocol and validity argument that establish cultural ambiguity as a property of items rather than asserting it. \textbf{(2)} We analyze MLLM-as-a-Judge alignment against culturally distinct reference pools, varying language, cultural persona, model origin, and reference-pool in-context demonstrations. \textbf{(3)} We argue that alignment under cultural ambiguity should be reported against multiple reference pools, with cross-pool divergence (which neither persona prompting nor reference-pool ICL removes) treated as a property of the judge rather than as noise.
\section{Related Work} \label{2_relatedwork}
\begin{figure*}[t]
        \centering       
        \includegraphics[width=.8\textwidth]{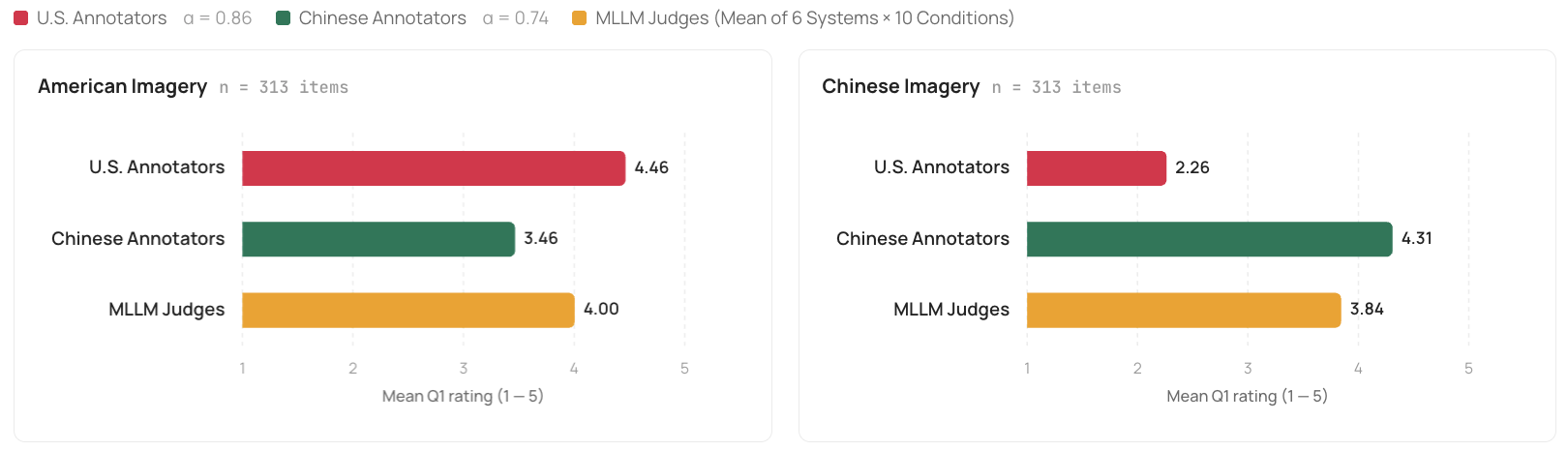}
    \caption{Mean Q1 Rating by Judge Type and Image Culture (1--5; $\alpha = \usnum{0.86}/\cnnum{0.74}$). The MLLM mean aligns with the U.S.\ consensus on American Imagery and with the Chinese consensus on Chinese Imagery.}
    \label{fig:fig2}
\end{figure*}

\subsection{LLMs and MLLMs as Judges}
Using language models as evaluators has become a common technique for tasks where reference-based metrics fall short. \citet{zheng2023judging} introduced the LLM-as-a-Judge framework with MT-Bench and Chatbot Arena, showing that strong models like GPT-4~\cite{openai2024gpt4technicalreport} approximate human pairwise preferences across conversational tasks. The paradigm has since been extended to tasks like summarization, instruction following, and long-form QA~\cite{fu-etal-2024-gptscore, pradeep-etal-2024-convkgyarn, dubois2023alpacafarm, min-etal-2023-factscore}, and is now embedded in major leaderboards~\cite{10.5555/3692070.3692401} and RLHF pipelines~\cite{10.5555/3692070.3693141}.

Multimodal extensions~\cite{10.5555/3692070.3692324} target outputs where metrics like CIDEr~\cite{Vedantam_2015_CVPR} and CLIPScore~\cite{hessel-etal-2021-clipscore} miss higher-level qualities such as grounding and instruction following. MLLM-as-a-Judge benchmarks frontier models on scoring, pairwise, and ranking tasks, while LLaVA-Critic~\cite{xiong2024llavacritic} and VLFeedback~\cite{li-etal-2024-vlfeedback} train open multimodal critics. A parallel line fine-tunes open judges on synthetic or human preference data, including Auto-J~\cite{li2024generative}, JudgeLM~\cite{zhu2025judgelm} and FARE~\cite{xu2026foundational}, and a third uses in-context demonstrations as a calibration mechanism for evaluator models. Across all three settings, judges are predominantly validated against aggregate human agreement, without specifying which population is being approximated.

\subsection{Cultural Bias in Multimodal Models}
Despite high aggregate agreement with humans, judges exhibit systematic biases including position~\cite{wang-etal-2024-large-language-models-fair}, length~\cite{panickssery2024llm}, self-preference~\cite{panickssery2024llm}, and stylistic sensitivity~\cite{chen-etal-2024-humans}. More broadly, language and multimodal models display social biases along gender, race, and political axes~\cite{gallegos-etal-2024-bias}, and reproduce stereotypes in generated images, captions, and visual reasoning~\cite{10.1145/3593013.3594095}.

Cultural bias is also well documented: LLMs disproportionately reflect WEIRD~\cite{335186}, English-language, and U.S.-centric perspectives~\cite{ramezani-xu-2023-knowledge}, with measurable gaps in cross-country opinion representation~\cite{10.5555/3618408.3619652} and inconsistent alignment across languages~\cite{alkhamissi-etal-2024-investigating}. Text-to-image systems underrepresent non-Western geographies and stereotype cultural concepts~\cite{basu2023inspectinggeographicalrepresentativenessimages, lee2025rewind}, while VLMs perform unevenly on culturally grounded VQA~\cite{romero2024cvqaculturallydiversemultilingualvisual}. Benchmarks such as CulturalBench~\cite{chiu-etal-2025-culturalbench} and BLEnD~\cite{myung2024blend} formalize this line of evaluation across varying cultural regions.

Most of this work examines what models produce rather than how they evaluate. A small but growing body of work probes judge behavior along non-cultural social axes and across languages~\cite{hada-etal-2024-metal}. We extend this inquiry to multimodal judgment in culturally ambiguous, preference-sensitive scenarios, comparing judges associated with different cultural contexts against culturally aligned human annotations.

\section{\benchmark{}} \label{3_benchmark}
We introduce \benchmark{}, a multimodal benchmark that quantifies the divergence between MLLM-as-a-Judge alignment and culturally distinct human reference pools under preference-sensitive ambiguity. Each item pairs an image with an evaluation prompt whose appropriate response hinges not on verifiable ground truth but on culturally situated evaluative expectations widely shared within one cultural context but not the other. We evaluate MLLMs under various conditions (Section~\ref{modelsandprompts}) and benchmark their judgments against culturally matched annotations from independent U.S.\ and Chinese annotator pools.
      
\subsection{Benchmark Construction} \label{3.1_benchmarkconstruction}
\subsubsection{Benchmark Composition}
\benchmark{} comprises 626 items balanced between U.S.- and mainland-Chinese-aligned concepts (313 each), spanning two cultural contexts each represented by frontier MLLMs developed for a distinct linguistic and regional market (Section~\ref{modelsandprompts}). Within each context, items span three categories in culturally specific cues~\cite{kim-etal-2025-tom, nayak-etal-2024-benchmarking}: food and beverage (147 per culture), fashion and human characteristics (64 per culture), and architecture and symbols (102 per culture). Items are concept-paired: each U.S.-aligned item has a mainland-Chinese counterpart with a matching referent (\eg{} U.S.\ and mainland-Chinese realizations of a celebratory meal; Figure~\ref{fig:fig1}).

\subsubsection{Image Sourcing}
All images in \benchmark{} are produced with \texttt{Gemini 3 Pro Image}~\citep{googledeepmind_nanobananapro_2026}. We use generated rather than naturalistic imagery because real photographs vary in aesthetic quality, lighting, composition, and learned human-preference signatures that influence evaluative ratings independently of the depicted subject~\citep{6247954,DBLP:journals/corr/KongSLMF16,kirstain2023pickapic,xu2023imagereward}. Each concept-paired generation uses matched parameters (same model, aspect ratio, lighting and composition register), so the only visual variable that differs across a pair is the cultural cue. Generator-imprinted signatures cannot drive the cross-pool divergence reported in Section~\ref{results}: such signatures would elicit shared rather than divergent responses across pools. A 30-pair naturalistic-photo replication reproduces the cross-pool Q1 divergence and the asymmetric judge bias at attenuated magnitudes (Appendix~\ref{real_image_verification}). We observe agreement on depicted content (Q2 placement at $\pm 1.5$ to $\pm 1.7$, or $76$--$85\%$ of the $\pm 2$ scale maximum) and disagreement on evaluation (cross-cultural Pearson $r = -0.12$ on Q1).

\subsubsection{Question Creation}
Prompts are authored by three bicultural expert annotators on a construction pool disjoint from the evaluation pool (Section~\ref{taskdesign_humanannotation}), under a rotating author/reviewer protocol. Iterative refinement ensures that prompts (i) elicit evaluative judgments rather than factual lookups, (ii) are plausibly contestable across cultures, and (iii) avoid surface-level stereotypes (Appendix~\ref{app:failure_modes}).

\begin{figure*}[t]
        \centering       
        \includegraphics[width=.8\textwidth]{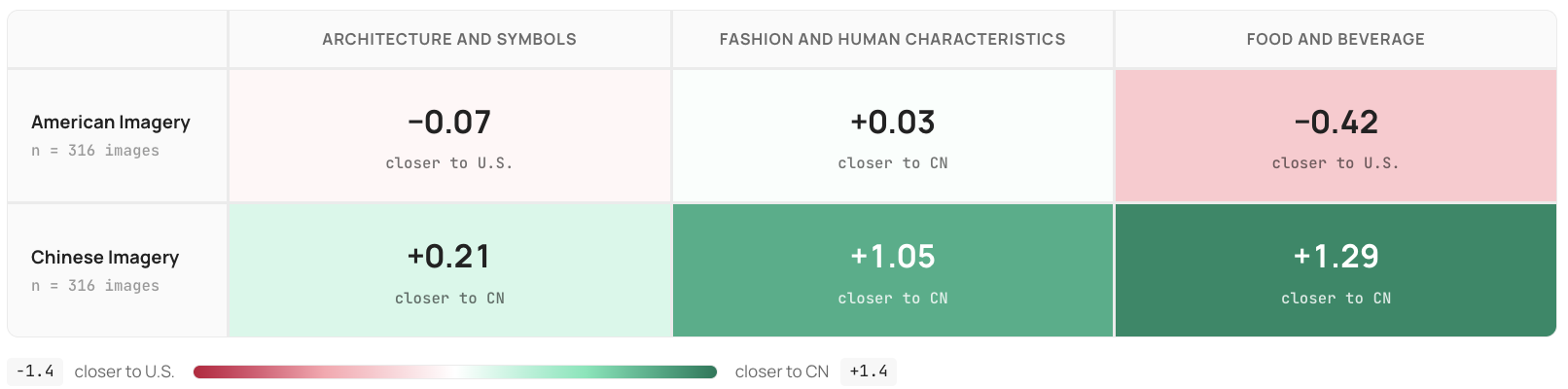}
    \caption{The cross-pool tilt $\Delta$ concentrates on Food and Beverage and on Fashion and Human Characteristics within Chinese Imagery; Architecture and Symbols is near-neutral.}
    \label{fig:fig3}
\end{figure*}

\subsubsection{Task Design and Human Annotation} \label{taskdesign_humanannotation}
Items in \benchmark{} were annotated by five U.S.\ and five mainland-Chinese bicultural-fluent annotators (ten total), compensated at \$20 per hour. The bicultural-fluency requirement makes Q2 cultural placement informative for both pools and likely attenuates the cross-pool Q1 divergence relative to a monocultural design. For each task, annotators answer four questions (Appendix~\ref{app:annotation-interface}):

\begin{enumerate}[label=\textbf{Q\arabic*.},leftmargin=2.4em,itemsep=2.5pt,topsep=4pt]
    \item \textbf{Evaluative response.} Respond to the prompt about the image on a 5-point ordinal scale.
    \item \textbf{Cultural alignment.} Rate the item on a 5-point Chinese to American scale.
    \item \textbf{Subjectivity.} Rate the Q1 prompt on a 5-point scale based on subjectivity.
    \item \textbf{Cross-cultural recognizability.} Rate on a 5-point scale how likely someone from the other cultural context is to understand the rationale behind the Q1 answer.
\end{enumerate}

\noindent Q4 proxies how widely a cultural reference has diffused across contexts, and how likely it might appear in training corpora. We use it to contextualize judge behavior across model origins.

\subsection{Models and Prompt Conditions} \label{modelsandprompts}
We evaluate six MLLMs: \texttt{claude-opus-4.7}, \texttt{gpt-5.5}, and \texttt{llama-4-scout} from U.S.\ developers; \texttt{glm-5}, \texttt{kimi-k2.6}, and \texttt{qwen-3-vl-flash} from mainland-Chinese developers~\cite{anthropic2026opus, openai2026gpt55, meta2025llama4, zai2026glm5, moonshot2026kimik26, qwen2026qwen3}. Each model rates every item under ten prompt conditions: two neutral (no persona; English and simplified Chinese) and eight persona conditions crossing persona detail (basic vs.\ advanced), persona culture (American vs.\ Chinese), and prompt language (English vs.\ Simplified Chinese), yielding $626 \times 6 \times 10 = 37{,}560$ judgment cells. Simplified Chinese prompts were translated with Gemini 3.1 Pro and verified by a bilingual member of the construction pool~\cite{gemini_3_1_pro}. Full prompt and model details are in Appendices~\ref{Appendix_Prompts} and~\ref{app:models:judges}.

\paragraph{In-context learning conditions.}
We additionally evaluate four in-context learning (ICL) conditions, formed by crossing demonstration annotator pool (U.S.\ vs.\ mainland-Chinese) with example imagery (American- vs.\ mainland-Chinese-aligned). Each ICL call prepends five (image, rating) demonstrations sampled one per Likert level 1--5 from a held-out pool disjoint from the 626 test items, then applies the baseline prompt. The four designs add $626 \times 6 \times 10 \times 4 = 150{,}240$ cells (187{,}800 total).

\subsection{Validating Cultural Ambiguity} \label{3.3}
\benchmark{}'s central validity claim is that items are within-pool reliable but cross-pool divergent in \emph{evaluation}, with the corpus splitting into anchors and contested probes used in Section~\ref{results} to separate cultural bias from item difficulty.

\paragraph{(a) Within-pool reliability.} Krippendorff's $\alpha$ on Q1 (ordinal distance) is $0.86$ (U.S.) and $0.74$ (Chinese), exceeding the $0.80$ and $0.667$ thresholds for definitive and tentative conclusions.

\paragraph{(b) Cross-pool disagreement.} Per-image Pearson on Q1 between U.S.\ and Chinese consensus is $r = -0.12$ (95\% CI $[-0.20, -0.04]$); mean absolute difference is $1.77$ on the 1--5 scale, reaching $2.89$ in the highest-divergence stratum (Chinese food items). The slight negative sign is informative: items rated higher by the U.S.\ pool are rated slightly lower by the Chinese pool, rather than the two pools being merely unrelated. Reliable within-pool agreement does not translate into cross-pool agreement. Because contested items are sampled to split the pools, on each one pool's reading is more permissive than the other's; the corpus-independent claim of this paper is the existence of an orientation residual after calibration is corrected (\S\ref{4.3}), while the sign of the residual on this corpus reflects which pool is the more-permissive on the contested majority of items.

\paragraph{(c) Disagreement is over evaluation, not recognition.} On Q2, American items are placed at $+1.61$ (U.S.) / $+1.70$ (Chinese) and Chinese items at $-1.51$ / $-1.68$---sign-identical, with the smallest absolute placement ($1.51$) at $76\%$ of scale maximum. Q3 confirms items as evaluative: $86\%$ (U.S.) and $88\%$ (Chinese) received Q3 $\leq 2$ (subjective on the 1--5 scale where 1 is most subjective). The cross-pool Q1 disagreement cannot be attributed to either pool failing to recognize content.

\paragraph{(d) Anchors and contested probes.} Ambiguity is graded rather than uniform: within-stratum cross-pool Pearson ranges from $0.04$ to $0.39$. Defining an item as a cultural anchor if both pool consensuses are $\geq 4$ (high anchor) or both $\leq 2$ (low anchor), and a contested probe otherwise, yields 227 high anchors, 39 low anchors, and 360 contested probes. Combining both within a single instrument is what allows Section~\ref{results} to attribute judge behavior to cultural bias rather than item difficulty.
\section{Results} \label{results}
We characterize how the six MLLM judges align with the U.S.\ and Chinese reference pools. Four claims structure the analysis: a positivity-floor calibration failure that, on this corpus, manifests as a Chinese-leaning lean (\S\ref{4.1}), a variance decomposition with a small origin tilt (\S\ref{4.2}), persona prompts that steer asymmetrically and leave a residual (\S\ref{4.3}), and reference-pool demonstrations that deepen rather than correct it (\S\ref{4.4}). For each model and condition we compute alignment with each pool (us-Pearson, us-MAE; cn-Pearson, cn-MAE) together with two tilt measures. The MAE-based tilt $\Delta = \text{us-MAE} - \text{cn-MAE}$ summarizes alignment-error asymmetry; positive values indicate that the judge fits the Chinese reference more closely. The normalized cultural tilt $T \in [-1, +1]$ summarizes placement on the ordinal rating scale itself, anchored at $T = -1$ for the Chinese human consensus and $T = +1$ for the U.S.\ consensus. $\Delta$ asks which pool the judge approximates with smaller error; $T$ asks where on the cultural axis the judge places the item.

\begin{figure*}[t]
        \centering       \includegraphics[width=.90\textwidth]{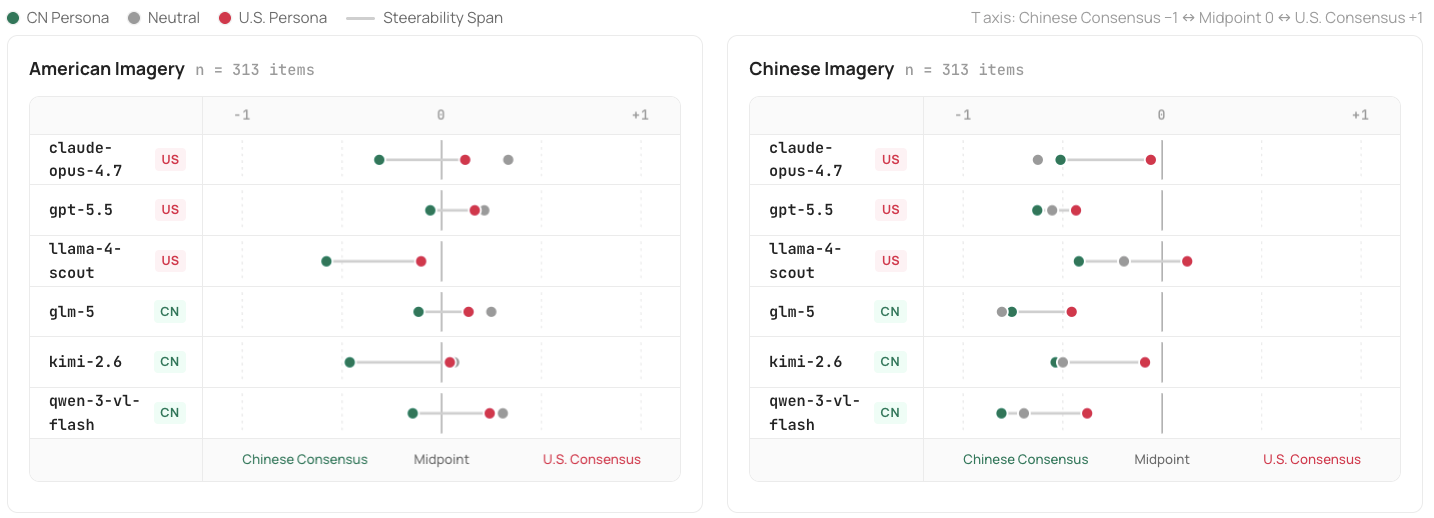}
    \caption{Persona prompts shift judges asymmetrically. Normalized cultural tilt $T$ per model under three persona conditions ($\cnnum{T=-1}$ Chinese consensus, $\usnum{T=+1}$ U.S.\ consensus). On Chinese imagery, five of six judges remain below the midpoint under every persona; only \texttt{llama-4-scout} crosses, reaching $T = +0.13$.}
    \label{fig:fig4}
\end{figure*}

\subsection{A Positivity Floor with Asymmetric Cultural Manifestation} \label{4.1}
\paragraph{Aggregate verdict, localized on Chinese imagery.}
Across the full set of judgment cells, mean us-MAE $= 1.367$ and mean cn-MAE $= 1.029$, yielding $\Delta = +0.339$ (paired Wilcoxon over images, $p = 4.5 \times 10^{-10}$). The model-mean $\Delta(\text{cn-Pearson} - \text{us-Pearson})$ is positive for every system. The aggregate tilt masks a sharp asymmetry across image culture: on American imagery, model error against U.S.\ humans is $0.87$ and against Chinese humans is $1.10$ (a slight U.S.\ lean); on Chinese imagery the comparison flips and widens, to $1.83$ and $0.94$ respectively. Figure~\ref{fig:fig2} contrasts the three juries, and Figure~\ref{fig:fig3} localizes the bias by image culture and category, showing that within Chinese imagery it concentrates on food and fashion rather than architecture.

\paragraph{Mechanism: a positivity floor.}
A rating-distribution view illuminates why the asymmetry exists. Models almost never produce 1s or 2s ($6\%$ to $16\%$ of rating cells across systems). On the $213$ items where the U.S.\ consensus is $\le 2$, only $10.0\%$ to $23.8\%$ of model cells fall that low; even the most aggressive U.S.-persona condition lifts this rate only to $22\%$ to $27\%$. The bias is therefore anchored to a calibration floor: models do not occupy the lower end of the ordinal range, and the floor they do produce ($\sim 3$) coincides mechanically with the more-permissive cultural reading.

\paragraph{Noise and cosmopolitan defaults.}
Two natural alternative accounts fail. First, the bias is not noise driven: per-image Pearson between $\Delta$ and the signed (U.S.\ minus CN) human gap is $-0.72$ ($p \approx 10^{-90}$, $n_{\text{images}} = 626$). Items the two pools agree on do not generate bias; items they disagree on do, and they do so by validating the more-permissive cultural reading. Second, judges are not retreating to globally-cosmopolitan defaults. Q4 establishes that Chinese references are systematically less recognizable cross-culturally (U.S.\ annotators rate $3.71/5$ in the symmetric direction; Chinese annotators rate $2.96/5$; paired Wilcoxon $p < 10^{-6}$), yet stratifying Chinese-imagery items by Q4 bin gives normalized tilt $T = -0.44$ at low recognizability and $T = -0.38$ at high recognizability. The bias is strongest where Western audiences are least likely to know the reference, ruling out the globally-recognized-default account.

\subsection{Image Dominates; Origin Adds a Tilt} \label{4.2}
To isolate where the bias lives in the experimental design, we regress per-cell $\Delta$ on item, model, condition, and origin terms. Image fixed effects dominate ($R^2 = 0.57$); model, condition, and origin contribute only marginal additional explanatory power (Table~\ref{table1}). Most of the variance is in the items, not the judges.

\paragraph{A small but present origin effect.}
Despite explaining little aggregate variance, model origin survives as a coefficient under image and condition fixed effects: the U.S.-origin coefficient on per-cell $\Delta$ is $-0.096$ ($p = 3.8 \times 10^{-7}$, item-clustered standard errors). Eight of ten conditions favor CN-origin and none are significantly reversed. On Chinese imagery, both U.S.- and CN-origin systems lean toward Chinese humans (the orientation residual of Section~\ref{4.1}), with U.S.-origin error against Chinese vs.\ U.S.\ humans at $1.00$ vs.\ $1.74$ and CN-origin at $0.88$ vs.\ $1.92$---both origins lean Chinese, with CN-origin leaning more. On American imagery, the lean is near-zero regardless of origin. Origin is thus a small additive tilt on top of a much larger calibration and orientation residual (robust in MAE space; not significant in Pearson space, $p = 0.52$), and the gap is approximately invariant under reference-pool ICL ($\pm 0.06$ in $T$-units of its baseline $+0.13$; Section~\ref{4.4}).

\subsection{Persona Prompts Leave a Residual} \label{4.3}
Across the ten prompt conditions, personas pull tilt in the expected direction (Figure~\ref{fig:fig4}): five of six models move further toward Chinese humans than toward Americans, and none reverses direction.

\paragraph{Steerability is bounded.}
Only one model crosses $T = 0$ on Chinese imagery (Figure~\ref{fig:fig4}); the five most-steerable models all show CN-pull $\ge$ US-pull. On Chinese food-and-beverage items where U.S.\ consensus is lowest, even the strongest U.S.-persona condition leaves model ratings more than a full ordinal point above the U.S.\ human floor. Within this bounded range, the dominant lever is the persona's cultural label: prompt-language variation ($|\Delta\text{Pearson}| < 0.06$) and persona detail level ($|\Delta| \approx 0.02$) contribute negligibly. ICL pushes in the opposite direction of the persona: under American-persona conditions on Chinese imagery, the persona-driven $T = -0.20$ collapses to $T = -0.30$ to $-0.37$ once any reference-pool ICL setting is added, and \texttt{llama-4-scout}'s only crossing of $T = 0$ in the baseline design ($T = +0.13$) does not survive any tested ICL design (Section~\ref{4.4}).

\paragraph{The residual asymmetry.}
Per-image-culture alignment is best when the persona matches the image culture: own-MAE drops from $1.09$ on mismatched personas to $0.81$ on matched ones. Yet after exhaustive per-target-culture optimization over models, conditions, and prompt languages, a cross-cultural asymmetry persists. The tightest fit obtainable on the Chinese-imagery $\to$ U.S.\ cell is $1.26$ MAE, against best same-pool fits of $0.59$ on American $\to$ U.S.\ and $0.68$ on Chinese $\to$ CN, and a best cross-pool counterpart of $0.96$ on American $\to$ CN.

\subsection{In-context Demonstrations Do Not Recalibrate}\label{4.4}
Persona prompts manipulate one elicitation surface; in-context demonstrations manipulate another. We tested whether reference-pool demonstrations close the residual by crossing demonstration annotator pool (U.S.\ vs.\ mainland-Chinese) with example imagery (American vs.\ Chinese), giving four ICL designs (\textsc{aa}, \textsc{ac}, \textsc{ca}, \textsc{cc}) evaluated alongside the ten baseline prompt conditions for all six judges ($n = 150{,}240$ ICL judgment cells).

\paragraph{Calibration is not restored.}
On Chinese imagery the share of cells rated 1 or 2 stays between 10.9\% and 12.7\% across all four ICL designs (vs.\ 12.7\% at baseline), while the share rated 5 rises by 4 to 7 percentage points; demonstrations widen the rating window upward, not downward. On items where the U.S.\ consensus is $\leq 2$ and imagery is Chinese, the floor-recovery rate is 14.2--16.6\% under ICL versus 16.5\% at baseline.

\paragraph{Orientation is deepened, not corrected.}
The Chinese-imagery normalized tilt $T$ moves from $-0.35$ (baseline) to between $-0.40$ and $-0.44$ across all four ICL designs. Paired Wilcoxon tests on per-image absolute error against the U.S.\ pool are significant for every design ($n = 18{,}780$ each; mean $\Delta|\text{err}|$ from $+0.04$ to $+0.09$; $p < 10^{-9}$ in every case, $p < 10^{-45}$ in three of four). The matched gain against the CN pool is of opposite sign and approximately equal magnitude, so per-target-pool ICL trades alignment with one pool for misalignment with the other in nearly equal magnitudes.

\paragraph{Source of demonstrations minimally matter.}
Holding example imagery fixed, switching from U.S.\ to Chinese demonstration annotators moves $T$ by $\leq 0.04$; holding annotator pool fixed, switching example imagery moves $T$ by $\leq 0.05$. Whose demonstrations a judge sees barely matters; the presence of any reference-pool demonstrations sharpens the same orientation residual. The single exception is \texttt{glm-5} under the \textsc{ac} design ($T -0.48 \rightarrow -0.39$), which is also the only judge whose origin and demonstration imagery cross. The best Chinese-imagery $\rightarrow$ U.S.-pool MAE per model averages 1.49 (baseline only) versus 1.48 (any configuration). The residual asymmetry of Section~\ref{4.3} is therefore unremovable by content-grounded calibration anchors as well as by content-free persona prompts.
\section{Findings and Discussion}
\label{sec:findings}

\begin{table*}[t]
    \centering\small
    \setlength{\tabcolsep}{8pt}
    \renewcommand{\arraystretch}{1.25}
    \begin{tabular}{@{}l w{r}{5.5em} w{r}{5.5em} @{\hspace{14pt}} w{r}{5.5em} w{r}{5.5em}@{}}
    \toprule
    &
      \multicolumn{2}{c}{\textbf{American
    Imagery}\hspace{0.4em}{\footnotesize\textcolor{vdgray}{$n{=}313$}}} &
      \multicolumn{2}{c}{\textbf{Chinese
    Imagery}\hspace{0.4em}{\footnotesize\textcolor{vdgray}{$n{=}313$}}} \\
    \cmidrule(lr){2-3} \cmidrule(lr){4-5}
    \textbf{Model}
      & {\textcolor{vdus}{$\to$ U.S.}} & {\textcolor{vdcn}{$\to$ CN}}
      & {\textcolor{vdus}{$\to$ U.S.}} & {\textcolor{vdcn}{$\to$ CN}} \\
    \midrule
    \texttt{claude-opus-4.7}~\originus
        & \vdcell{0.61}{EN/--}{44} & \vdcell{0.97}{EN/CN}{70}
        & \cellcolor{vdresid}\vdcellbest{1.26}{ZH/US}{58}
        & \vdcell{0.78}{EN/--}{36} \\
    \texttt{gpt-5.5}~\originus
        & \vdcell{0.65}{EN/US}{47}
        & \cellcolor{vdsame}\vdcellbest{0.96}{EN/CN}{70}
        & \vdcell{1.71}{EN/US}{79} & \vdcell{0.71}{EN/CN}{33} \\
    \texttt{llama-4-scout}~\originus
        & \vdcell{0.96}{ZH/US}{70} & \vdcell{1.04}{EN/CN}{75}
        & \vdcell{1.35}{EN/US}{63} & \vdcell{0.94}{ZH/CN}{44} \\
    \texttt{glm-5}~\origincn
        & \cellcolor{vdresid}\vdcellbest{0.59}{EN/--}{43}
        & \vdcell{0.96}{EN/CN}{70}
        & \vdcell{1.69}{EN/US}{78}
        & \cellcolor{vdsame}\vdcellbest{0.68}{EN/--}{31} \\
    \texttt{kimi-k2.6}~\origincn
        & \vdcell{0.78}{EN/--}{57} & \vdcell{0.99}{EN/CN}{72}
        & \vdcell{1.37}{EN/US}{63} & \vdcell{0.79}{ZH/CN}{37} \\
    \texttt{qwen-3-vl-flash}~\origincn
        & \vdcell{0.66}{EN/US}{48} & \vdcell{1.06}{EN/CN}{77}
        & \vdcell{1.54}{EN/US}{71} & \vdcell{0.68}{ZH/CN}{31} \\
    \midrule
    \textit{Inter-cultural Human MAE}
        & \multicolumn{2}{c}{\textit{\textcolor{vdgray}{1.38}}}
        & \multicolumn{2}{c}{\textit{\textcolor{vdgray}{2.16}}} \\
    \bottomrule
    \end{tabular}
    \caption{Best MAE per (Image Culture, Target Pool) over 10 Conditions; cells show MAE / Condition /
    [\%] of human MAE. Column floors highlighted: \colorbox{vdresid}{\strut\,$\to$ U.S.\,},
    \colorbox{vdsame}{\strut\,$\to$ CN\,}.}
    \label{table1}
  \end{table*}

\subsection{Alignment is a Profile, Not a Scalar}
\label{sec:findings-profile}

\benchmark{} items are within-pool reliable but cross-pool divergent (Section~\ref{3.3}), so any single ``agreement with humans'' number silently conditions on a cultural reference pool before the score is computed. A judge with us-Pearson $0.50$ and cn-Pearson $0.10$ is not ``a good judge of human preference''; it is a U.S.-leaning judge that performs poorly on the Chinese reference. Best-model-to-human quadratic-weighted $\kappa$ reaches $0.27$ on the better-aligned pool while inter-cultural human-to-human $\kappa$ is below zero: models exceed inter-cultural human agreement by silently picking the Chinese side, not by capturing both. We therefore recommend (R1, below) that alignment be reported against each reference pool, with divergence treated as a property of the judge rather than absorbed as noise. The point generalizes to any preference-sensitive evaluation with a heterogeneous human pool: regional dialects, age cohorts, expertise levels, communities of practice~\citep{pavlick-kwiatkowski-2019-inherent}.

\subsection{Calibration and Orientation}
\label{sec:findings-decomp}
Models register a faint cultural signal in the human-correct direction: the model raw rating gap (American minus Chinese imagery) is $+0.16$, sign-matched to U.S.\ humans ($+2.20$) and opposite to CN humans ($-0.85$). But the magnitude is compressed, so the practical effect is near-flat treatment of both image cultures. That compression and the residual sign are, respectively, the calibration and orientation story.

Calibration failure is the inability to produce low-end ratings (Section~\ref{results}). Orientation failure is what remains after calibration is partially recovered: even under U.S.\ persona, the residual sign of the cultural tilt $T$ stays negative on Chinese imagery.

Persona prompts touch orientation (they reweight which cultural reading the judge defaults to) but do not, on their own, fix calibration (they do not teach the judge that a Chinese-aligned item rated by U.S.\ taste lives at the bottom of the ordinal scale). De-biasing must therefore address both failures, and Section~\ref{4.4} shows that the most natural candidate---pairing prompt elicitation with reference-pool calibration anchors---does not: demonstrations widen the rating window upward without restoring use of the low end, deepen the orientation tilt, and trade alignment with one pool for misalignment with the other in approximately equal magnitudes. Calibration and orientation therefore appear to be properties of the judges' rating distributions, not of the tested regimes.

\subsection{Origin is not Competence}
\label{sec:findings-origin}
The $-0.096$ origin coefficient documented in Section~\ref{4.2} places the origin effect in a $\sim 0.10$ MAE band, well below the $\sim 0.5$ MAE band of persona effects, which itself sits well below the $\sim 1$ MAE inter-cultural human gap. The hierarchy is one-directional: choosing model origin matters less than choosing prompt configuration, and choosing prompt configuration matters less than recognizing inherent cultural multiplicity in the items.

The shape of the origin effect is also informative. A pure-competence account (CN-origin models simply know Chinese culture better) predicts origin-conditional symmetry: CN-origin matches Chinese humans better, U.S.-origin matches American humans better. We do not observe symmetry. On Chinese imagery, all six systems lean toward Chinese humans regardless of origin (Section~\ref{4.1}); the competence account predicts a crossed pattern that does not appear. We therefore treat origin as a covariate worth reporting alongside model identity (R2, below), not as a primary lever for de-biasing.

\subsection{Recommendations}
\label{sec:findings-rec}
\paragraph{(R1) Multi-pool reporting.} When items are preference-sensitive or culturally loaded, report alignment against each available reference pool and treat divergence between those numbers as a property of the judge. A single Pearson or MAE against a single pool is, on these items, a category error. The recommendation holds even when the judge is conditioned on reference-pool demonstrations: Section~\ref{4.4} shows that target-pool ICL can be made arbitrarily favorable to one pool without the underlying judge changing.

\paragraph{(R2) Origin as covariate.} When comparing judges by ``human alignment'' on culturally loaded benchmarks, treat model origin as a covariate to report alongside model identity. Origin-blind comparisons systematically advantage one cultural alignment over the other (Section \ref{sec:findings-origin}). The effect is small ($\sim 0.10$ MAE) but robust under image and condition fixed effects; not disclosing it allows leaderboards to inherit cultural lean silently.
\section{Conclusion}
We introduced \benchmark{}, a multimodal benchmark that measures how MLLM-as-Judge alignment splits across culturally distinct human reference pools under preference-sensitive ambiguity. We find two failures: a positivity-floor calibration failure (models refuse to produce low-end ratings) and an orientation failure (default to one cultural norm). On this corpus, where contested items split the two pools, the floor mechanically validates the more-permissive Chinese reading; persona prompting partially recovers calibration, but the orientation residual survives. Neither persona prompting nor reference-pool in-context demonstrations close it. These findings motivate a methodological shift: when evaluation items are culturally loaded, report alignment against each reference pool separately and treat divergence as judge-level information about which cultural norm the model encodes.
\newpage
\section*{Limitations}

\paragraph{Scope.}
We restrict the cultural investigation to U.S.\ and mainland Chinese evaluative norms across three visual categories (food and beverage, fashion and human characteristics, architecture and symbols). ``American'' and ``mainland Chinese'' function as construct labels for the annotator pools, not as claims about monolithic national groups. Extension to additional cultural pairings and categories is left to future work.

\paragraph{Annotator pool and stimulus type.}
Reference judgments come from ten bicultural-fluent expert annotators (five per culture). All images are produced with Gemini 3 Pro Image; a 30-pair naturalistic-photography probe (Appendix~\ref{real_image_verification}) reproduces the cross-pool divergence at attenuated magnitudes.

\paragraph{Model origin as proxy.}
We group models by developer geography and treat this as a proxy for sociotechnical factors (training-data composition, alignment-annotator demographics, preference-modeling pipelines). We do not have access to training corpora or RLHF annotator pools of any evaluated system and cannot attribute the origin effect to any single factor. We frame origin as a covariate worth disclosing, not as a mechanistic explanation.

\paragraph{Purpose of work.}
The benchmark is descriptive, not normative: both annotator pools are reliable within-pool, and the cross-cultural gap is itself the finding. It should not be used to rank cultural perspectives by correctness, select evaluator models on the basis of cultural alignment as a proxy for quality, or exclude minority-pool perspectives on the grounds of divergence.

% \section*{Acknowledgments}
% thank you to revanth gangi reddy and sridhar raghavan for feedback. angela wang and jean li for supporting annotations.

\bibliography{custom}

\newpage
\appendix

\section{Verifying with Real Images}
\label{real_image_verification}
To verify that this divergence is not an item of generated imagery, we collected a 30-pair ablation (60 photographs) of naturalistic U.S. and mainland Chinese images for concepts that already appear in \benchmark, ran the same six MLLM judges under all ten prompt conditions, and
re-annotated Q1--Q4 in each cultural pool with a five-annotator consensus following the protocol of Section~\ref{3_benchmark}. On the matched 60-image subset, the cross-pool Pearson on Q1 is $r = 0.06$ for real photographs and $r = 0.06$ for the synthetic counterparts (full-corpus reference: $r = -0.12$). Per-item signed cross-pool gaps are highly correlated across regimes ($r = 0.72$), and a Wilcoxon paired test on the $|\text{U.S.} - \text{CN}|$ gap finds no significant difference between regimes ($z = -1.19$, $p = 0.24$). Both pools continue to agree on Q2 cultural placement on real photos (American items at $+1.67$ / $+1.80$; Chinese items at $-1.60$ / $-1.87$; sign concordant on $80\%$ of items), confirming that the Q1 disagreement is not driven by either pool failing to recognize the depicted culture. The asymmetric judge bias central to Section~\ref{results} (tilt toward Chinese humans on Chinese items with no symmetric tilt on American items) also reproduces on real photographs: across the $60$ (model $\times$ condition) cells, mean tilt-to-CN is $+0.18$ on Chinese items and $-0.39$ on American items, with the sign of the asymmetry preserved in $50$ of $60$ cells and a cell-level Pearson of $+0.82$ between the real and synthetic asymmetry vectors. Magnitudes are mildly attenuated on real photos (cross-pool MAE $1.53$ vs.\ $1.72$; tilt-to-CN on Chinese items $+0.18$ vs.\ $+0.44$), in the direction predicted by the present argument: naturalistic photos add aesthetic and compositional variance orthogonal to the cultural cue. We treat this ablation as a validity probe at $n = 30$ concept pairs rather than a powered replication; the relevant claim is qualitative preservation of the cross-pool divergence and the asymmetric judge bias on real photographs of the matched concepts.

\section{Human Annotation} \label{appendix_human_annotation}
We recruited 10 annotators using three criteria intended to capture sustained engagement with the target culture: (i) self-identified cultural and ethnic affiliation with the target group; (ii) regular engagement with that country's social media (10+ hours a week), as a proxy for ongoing cultural participation; and (iii) fluency in the target language(s). All annotators are between 24 and 30 years old to control for generational variation in cultural reference points. Annotators were recruited through university-affiliated social media groups and compensated at \$20 per hour, a rate exceeding the local government-mandated minimum wage. All annotators were informed that their ratings would be used in a research paper.

\section{Dataset Release}
The dataset will be released under CC BY 4.0.

\section{Human Annotation Interface}
\label{app:annotation-interface}
\begin{figure*}[t]
  \centering
  \includegraphics[width=0.8\linewidth]{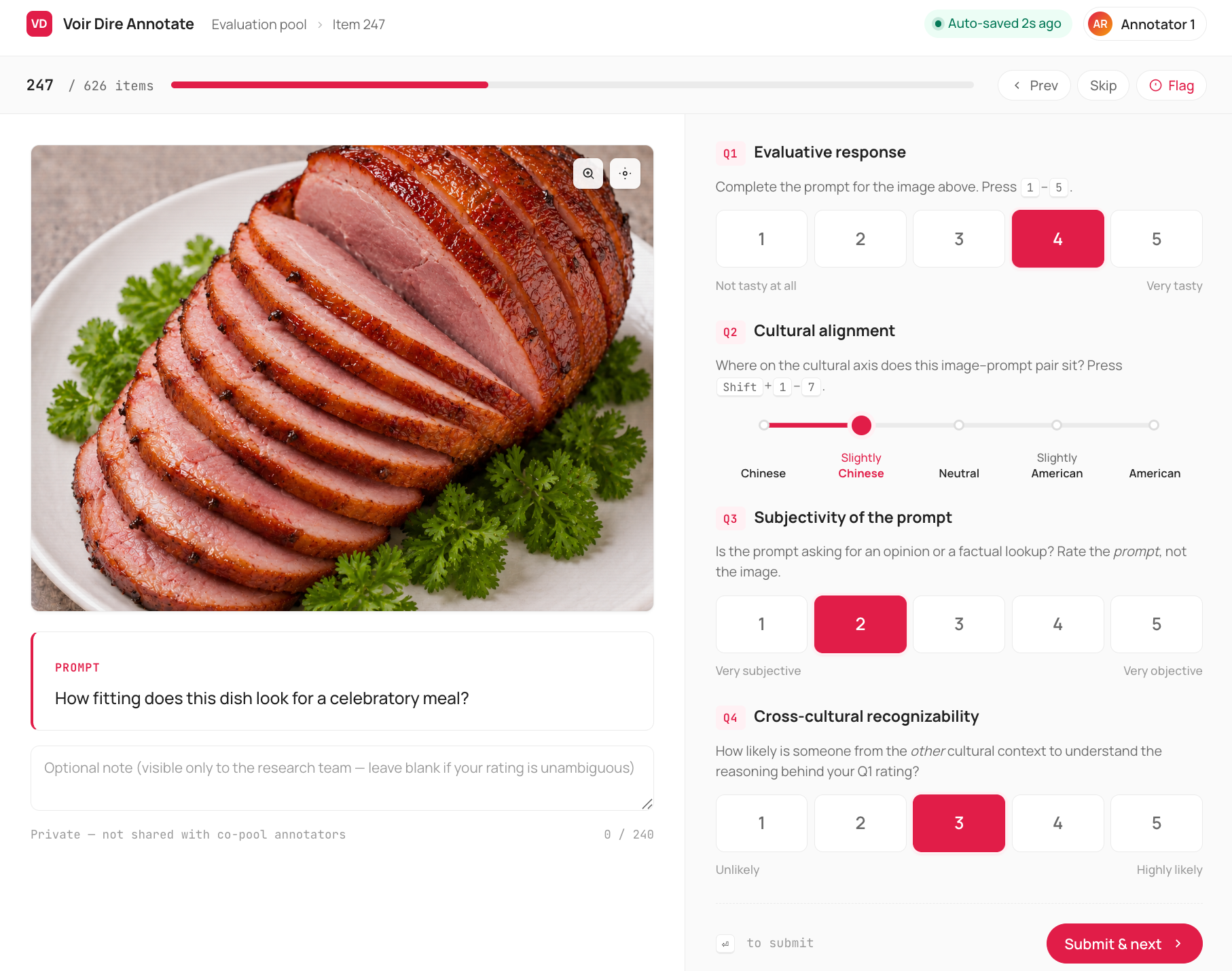}
  \caption{The human annotation interface, shown for a U.S.\ food item. Annotators see the image and prompt on the left and answer Q1--Q4 on the right; the Q2 cultural-placement scale uses a five-point slider with anchored labels (\textit{Chinese} through \textit{American}).}
  \label{fig:annotation-interface}
\end{figure*}

Figure~\ref{fig:annotation-interface} shows the interface that each annotator used while rating items.

\section{Models}
\label{app:models:judges}

\subsection{Image Generation}
\label{app:models:imggen}
All 626 paired stimuli were generated with Google's \texttt{gemini-3-pro-image-preview} model via the \texttt{google-genai} Python SDK. Generation was performed in a single batch on \textbf{April 18, 2026}.

\begin{table}[h]
\centering
\small
\setlength{\tabcolsep}{6pt}
\renewcommand{\arraystretch}{1.25}
\begin{tabular}{@{}>{\bfseries}lp{0.62\linewidth}@{}}
\toprule
Model & \texttt{gemini-3-pro-image-preview} \\
Provider & Google \\
SDK / call & \texttt{google-genai} \textperiodcentered{} \texttt{client.models.generate\_content} \\
Modality & \texttt{IMAGE} (image-only response) \\
Aspect ratio & $1{:}1$ \\
Image size & \texttt{1K} \\
Access date & April 18, 2026 (single batch) \\
\bottomrule
\end{tabular}
\caption{Image generation configuration. Identical parameters across all U.S./Chinese concept pairs hold photographic confounds fixed.}
\label{tab:imggen}
\end{table}

\subsection{Judge Models}
We evaluate six MLLMs spanning three capability tiers and two model-origin contexts (three U.S.\ and three mainland-Chinese). Every model is queried under all ten prompt conditions on every item, yielding $626 \times 6 \times 10 = 37{,}560$ baseline judgment cells. The four in-context demonstration conditions of Section~\ref{4.4} add a further $626 \times 6 \times 10 \times 4 = 150{,}240$ cells. The baseline evaluation ran from April~18 through April~25, 2026; the demonstration conditions ran May 16--18, 2026.

\begin{table*}[h]
\centering
\small
\setlength{\tabcolsep}{4pt}
\renewcommand{\arraystretch}{1.2}
\resizebox{0.96\linewidth}{!}{%
\begin{tabular}{@{}llllc@{}}
\toprule
\textbf{Model} & \textbf{Provider} & \textbf{API model ID} & \textbf{Endpoint / SDK} & \textbf{Mode} \\
\midrule
\multicolumn{5}{@{}l}{\textit{Image Generation Model}} \\
\midrule
Gemini 3 Pro Image & Google          & \texttt{gemini-3-pro-image-preview} & \texttt{google-genai} SDK (\texttt{generate\_content}) & Image \\
\midrule
\multicolumn{5}{@{}l}{\textit{Thinking tier}} \\
\midrule
GPT-5.5         & OpenAI (US)     & \texttt{gpt-5.5}                   & \texttt{openai} SDK (default endpoint)             & Thinking \\
Kimi K2.6       & Moonshot (CN)   & \texttt{kimi-k2.6}                 & \texttt{api.moonshot.ai/v1} (OpenAI-compatible)    & Thinking \\
\midrule
\multicolumn{5}{@{}l}{\textit{Non-thinking tier, large}} \\
\midrule
Claude Opus 4.7 & Anthropic (US)  & \texttt{claude-opus-4-7}           & \texttt{anthropic} SDK (default endpoint)          & Direct \\
GLM-5V-Turbo    & Z.AI (CN)       & \texttt{glm-5v-turbo}              & \texttt{zai-sdk} (native \texttt{ZaiClient})       & Direct \\
\midrule
\multicolumn{5}{@{}l}{\textit{Non-thinking tier, smaller}} \\
\midrule
Llama 4 Scout   & Meta (US)       & \texttt{meta-llama/llama-4-scout}  & \texttt{openrouter.ai/api/v1}                  & Direct \\
Qwen-3-VL-Flash    & Alibaba (CN)    & \texttt{qwen3-vl-flash}            & \texttt{dashscope-intl.aliyuncs.com/compatible-mode/v1} & Direct \\
\bottomrule
\end{tabular}%
}
\caption{Model providers, API identifiers, endpoints, and operating mode. Total usage of these models cost approximately \$1700 USD.}
\label{tab:judges}
\end{table*}

\paragraph{Strict output parsing.}
Each judge call requires a single integer rating in $\{1,2,3,4,5\}$ as the entire reply. We apply strict pattern matching to ensure consistency and, on a malformed reply, retry up to five times while progressively tightening the instruction.

\section{Prompts} \label{Appendix_Prompts}

Each judgment cell comprises three fields: an optional persona prefix, the item's evaluative question, and a language-matched rating instruction.

\[
\underbrace{\text{[Persona prefix]}}_{\text{optional}}
\;+\;
\underbrace{\text{\{question\}}}_{\text{per item}}
\;+\;
\underbrace{\text{[Rating prompt]}}_{\text{language-matched}}
\]

\subsection{Rating Instructions}
\label{app:prompts:rating}

\begin{tcolorbox}[promptbox, title={Rating instruction}]
\textbf{English.}\\[2pt]
Respond with ONLY a single integer from 1 to 5. No words, no punctuation, no explanation --- just the digit.
\tcblower
\begin{cn}
\textbf{简体中文.}\\[2pt]
请仅回复一个 1 到 5 之间的整数。不要包含任何其他文字、标点或解释 —— 只回复一个数字。
\end{cn}
\end{tcolorbox}

\subsection{Neutral Prompts}
\label{app:prompts:neutral}

\begin{tcolorbox}[promptbox, title={Neutral prompt}]
\textbf{English.}\\[2pt]
\textit{\{english\_question\}}\\[3pt]
Respond with ONLY a single integer from 1 to 5. No words, no punctuation, no explanation --- just the digit.
\tcblower
\begin{cn}
\textbf{简体中文.}\\[2pt]
\textit{\{chinese\_question\}}\\[3pt]
你的回复只能是一个 1 到 5 之间的整数。不能使用任何文字、标点符号 —— 不需要解释只需要回复数字本身。
\end{cn}
\end{tcolorbox}

\subsection{Basic Persona Prompts}
\label{app:prompts:basic}

\begin{tcolorbox}[promptbox, title={Basic persona: American}]
\textbf{English.}\\[2pt]
You are an American. Respond from the perspective of a typical American person, drawing on American cultural values, norms, and intuitions.
\tcblower
\begin{cn}
\textbf{简体中文.}\\[2pt]
你是一名美国人。作为一个典型美国人，请结合美国社会中普遍认同的价值取向、行为规范和判断方式来回应。
\end{cn}
\end{tcolorbox}

\vspace{6pt}

\begin{tcolorbox}[promptbox, title={Basic persona: Chinese}]
\textbf{English.}\\[2pt]
You are a Chinese person. Respond from the perspective of a typical Chinese person, drawing on Chinese cultural values, norms, and intuitions.
\tcblower
\begin{cn}
\textbf{简体中文.}\\[2pt]
你是一名中国人。作为一个典型中国人，请结合中国社会中普遍认同的价值取向、行为规范和判断方式来回应。
\end{cn}
\end{tcolorbox}

\subsection{Advanced Persona Prompts}
\label{app:prompts:advanced}

\begin{tcolorbox}[promptbox, title={Advanced persona: American}]
\textbf{English.}\\[2pt]
You self-identify as a part of American culture. You are between 24 and 30 years old. You regularly engage with American social media (Instagram, Snapchat, and TikTok) and are fluent in English.
\tcblower
\begin{cn}
\textbf{简体中文.}\\[2pt]
你自我认同为美国文化的一部分。你的年龄介于 24 至 30 岁之间。你经常使用美国主流社交媒体平台（Instagram、Snapchat 和 TikTok），并且英语流利。
\end{cn}
\end{tcolorbox}

\vspace{6pt}

\begin{tcolorbox}[promptbox, title={Advanced persona: Chinese}]
\textbf{English.}\\[2pt]
You self-identify as a part of Chinese culture. You are between 24 and 30 years old. You regularly engage with Chinese social media (WeChat, Douyin, and Xiaohongshu) and are fluent in simplified Chinese.
\tcblower
\begin{cn}
\textbf{简体中文.}\\[2pt]
你自我认同为中国文化的一部分。你的年龄介于 24 至 30 岁之间。你经常使用中国主流社交媒体平台（微信、抖音和小红书），并且能流利地使用简体中文。
\end{cn}
\end{tcolorbox}

\section{Authored Prompt-Image Failures} \label{app:failure_modes}

\begin{figure*}[t]
  \centering
  \includegraphics[width=0.96\linewidth]{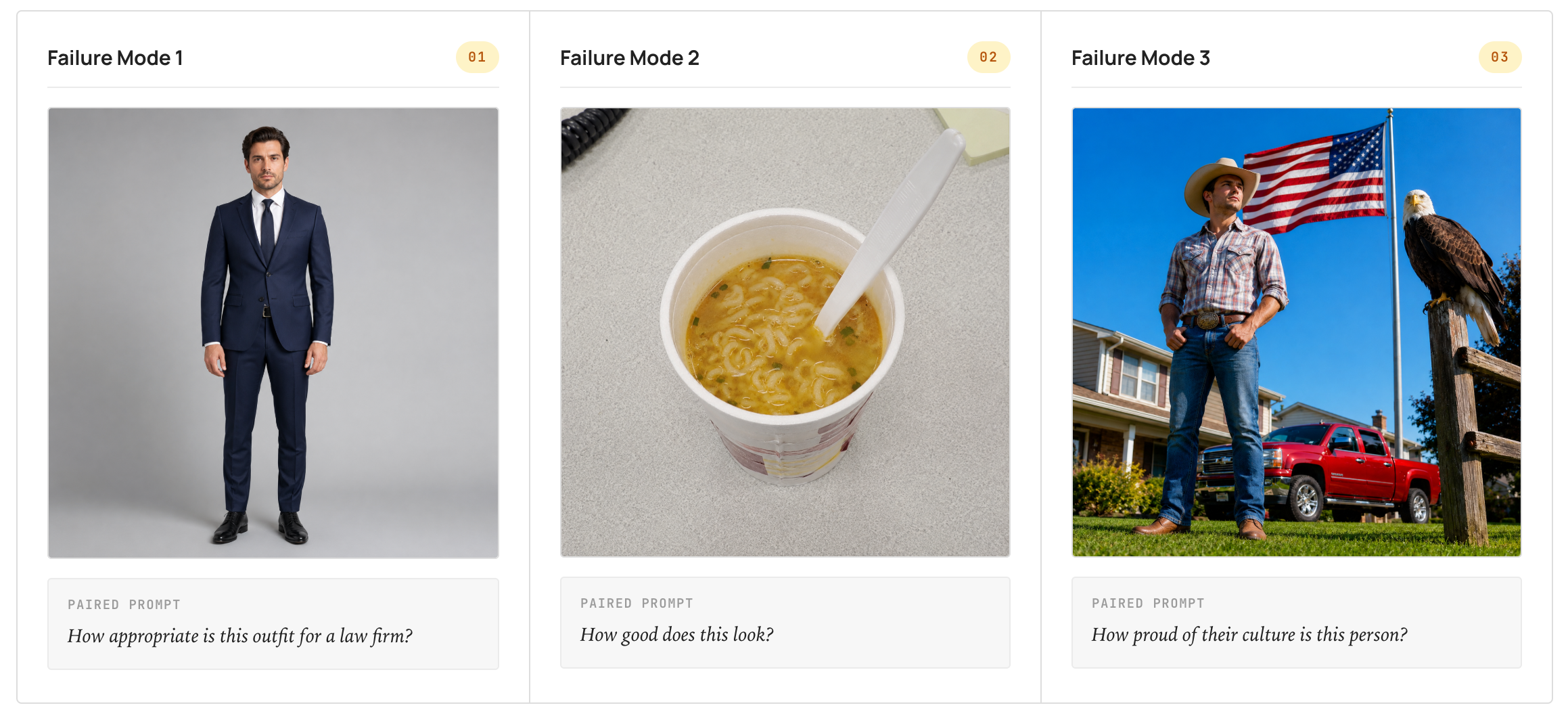}
  \caption{Three failure modes encountered while authoring prompt--image pairs. \emph{Left:} the prompt has a factual answer rather than eliciting an evaluative judgment. \emph{Middle:} the prompt is too generic to be culturally contestable, and both pools converge. \emph{Right:} the image is saturated with stereotypical cultural iconography, collapsing the prompt into a tautology. Pairs of this kind were excluded from the released benchmark.}
  \label{fig:failure-modes}
\end{figure*}

During stimulus construction we iteratively pruned prompt--image pairs that did not satisfy the design requirements laid out in Section~\ref{3_benchmark}: the pair must (i) elicit a genuinely \emph{evaluative} judgment on a $1$--$5$ scale rather than a factual or definitional one, (ii) be plausibly \emph{contestable across cultures}, so that the U.S.\ and Chinese annotator pools have grounds to diverge, and (iii) probe culturally contingent norms rather than ostentatious cultural iconography. One representative pair for each is illustrated in Figure~\ref{fig:failure-modes}.

\paragraph{Failure Mode 1: the prompt does not elicit an evaluative judgment.}
The left panel of Figure~\ref{fig:failure-modes} pairs a tailored navy suit with \emph{``How appropriate is this outfit for a law firm?''} A dark suit is the canonical law-firm garment in both U.S.\ and Chinese white-collar contexts, so the answer is equal regardless of culture; pilot ratings piled at the ceiling in both pools. The pair fails requirement~(i): it probes \emph{factual} convention rather than \emph{evaluative} judgment. We exclude pairs whose answer is recoverable from a dress-code manual.

\paragraph{Failure Mode 2: the answer is not plausibly contestable across cultures.}
The middle panel pairs an instant-noodle cup with \emph{``How good does this look?''} Instant ramen is globally diffused, sold in near-identical formats and carrying the same low-prestige connotation in both pools, so the two have no grounds to diverge, and pilot ratings converged on a mid-high score. We exclude items whose referent has been absorbed into both cultures under a single de facto norm, preferring concept pairs whose realizations remain culturally distinctive.

\paragraph{Failure Mode 3: cultural cues reduce to surface-level stereotypes.}
The right panel pairs a man with a U.S.\ flag, bald eagle, red pickup, and suburban houses with \emph{``How proud of their culture is this person?''} The image is composed almost entirely of stock shorthand for ``America,'' so the prompt asks annotators to evaluate what the image is reading off prop density rather than making a normative judgment. The pair fails requirement~(iii): the construct is carried by \emph{iconography} rather than \emph{norms}. We exclude stimuli built on saturated symbolic props, requiring lower-iconicity cues (everyday clothing, food presentation, posture, domestic setting) that leave room for divergent normative readings.

\section{In-Context Demonstration Conditions}\label{app:ICL}
\subsection{Design}
Each in-context demonstration cell is defined by (i) the demonstration annotator pool, U.S.\ or mainland-Chinese, and (ii) the demonstration image culture, American-aligned or Chinese-aligned, yielding four designs labeled \textsc{aa}, \textsc{ac}, \textsc{ca}, \textsc{cc} (annotator--image). Each call prepends 5 (image, rating) demonstrations to the standard prompt template, then queries the judge under the same parsing protocol. Demonstrations are sampled with one demonstration at each Likert rating level 1--5 and from a held-out demonstration pool disjoint from the 626 test items; the same demonstration set is used across the six judges within a (design, prompt-condition, item) cell to isolate judge-side variance. We hold demonstration count, sampling protocol, and instruction text fixed across the four designs so the only experimental factors are the cultural source of the demonstrations and the image culture they were drawn from.

\subsection{Rating Distributions}
The 1- or 2-rating share on Chinese imagery is 10.9\% to 12.7\% under ICL versus 12.7\% at baseline; the 5-rating share rises by 4--7 percentage points. The mean rating on Chinese imagery rises from 3.84 to between 3.90 and 3.98.

\subsection{Best per-target-pool MAE under the full design}
The Chinese-imagery $\rightarrow$ U.S.-pool best-MAE per-model average is 1.486 (baseline) versus 1.480 (any configuration), an improvement of 0.4\%. No model improves the Chinese-imagery $\rightarrow$ U.S.-pool fit by more than 0.32 MAE under any ICL design.

\subsection{Per-model Normalized Cultural Tilt $T$ under ICL}
The five most CN-leaning judges all become more CN-leaning under at least three of the four designs; \texttt{llama-4-scout}, the only judge whose persona-conditioned $T$ crosses zero on Chinese imagery in the baseline design, does not cross zero under any ICL design (max $T = -0.36$). \texttt{glm-5} under \textsc{ac} is the only (model, design) cell where $T$ moves toward zero by more than $0.05$ ($T -0.48 \rightarrow -0.39$).

\subsection{Persona $\times$ ICL Interaction}
On Chinese imagery, restricted to American-persona conditions only, baseline $T = -0.20$ collapses to $T = -0.33$ (\textsc{aa}), $-0.30$ (\textsc{ac}), $-0.30$ (\textsc{ca}), $-0.37$ (\textsc{cc}). Demonstrations erase roughly half of the persona-driven correction toward the U.S.\ pool.

\subsection{Origin $\times$ ICL}
The origin gap on Chinese imagery (US-origin $T$ minus CN-origin $T$) is $+0.13$ at baseline and $+0.11$, $+0.10$, $+0.14$, $+0.07$ under \textsc{aa}, \textsc{ac}, \textsc{ca}, \textsc{cc} respectively. The gap is preserved under ICL within $\pm 0.06$.

\subsection{Category Breakdown}
The food $\times$ Chinese imagery cell is the worst at baseline ($\Delta = +1.29$) and the worst under all four ICL designs ($\Delta$ ranges from $+1.36$ to $+1.47$). The architecture $\times$ Chinese imagery cell---near-neutral at baseline ($\Delta = +0.21$)---gains the largest absolute increase under ICL ($\Delta$ ranges from $+0.38$ to $+0.47$) the demonstrations convert ambiguity into commitment in the same direction the model already leans.

\subsection{Statistical Tests}
Per-image absolute-error paired Wilcoxon tests on Chinese imagery ($n = 18{,}780$ per cell) yield $p < 10^{-9}$ for every (ICL design $\times$ target pool) combination, with mean differences of opposite sign for the U.S.\ and CN pools and approximately equal magnitudes (e.g., \textsc{aa}: U.S.\ mean $\Delta|\text{err}| = +0.091$, CN mean $\Delta|\text{err}| = -0.094$).

\end{document}